\pdfoutput=1
\documentclass[11pt]{article}

\usepackage[final]{acl}
\usepackage{natbib}
\usepackage{times}
\usepackage{subcaption}
\usepackage{latexsym}
\usepackage{booktabs}
\usepackage{fancyvrb}
\usepackage{enumitem}
\usepackage{multirow}
\usepackage{float}
\usepackage[T1]{fontenc}
\usepackage[utf8]{inputenc}
\usepackage{adjustbox}

\usepackage{microtype}

\usepackage{inconsolata}

\usepackage{graphicx}
\usepackage{xcolor}  
\usepackage[utf8]{inputenc} % allow utf-8 input
\usepackage[T1]{fontenc}    % use 8-bit T1 fonts
\usepackage{hyperref}       % hyperlinks
\usepackage{url}            % simple URL typesetting
\usepackage{amsfonts}       % blackboard math symbols
\usepackage{nicefrac}       % compact symbols for 1/2, etc.
\usepackage{microtype}      % microtypography
\usepackage{graphicx}
\usepackage[most]{tcolorbox}
\usepackage{colortbl}
\usepackage{listings}

\lstdefinestyle{pythonstyle}{
    backgroundcolor=\color{gray!15},   % Set background color
    commentstyle=\color{blue},          % Set comment color
    keywordstyle=\color{magenta},         % Set keyword color
    numberstyle=\tiny\color{gray},       % Set line number style
    numbers=left,                        % Position of line numbers
    stepnumber=1,                        % Step size for line numbers
    numbersep=5pt,                       % Space between line numbers and code
    stringstyle=\color{red},             % Set string color
    basicstyle=\ttfamily\footnotesize,   % Basic font style
    breaklines=true,                     % Automatically break lines
    captionpos=b,                        % Position of the caption
    escapeinside={\%*}{*)}               % Escape inside comments
}

\usepackage[linesnumbered,ruled,vlined]{algorithm2e}

\title{\method{}: An Automated Framework for Evolving Agentic Workflows}

\author{
Yingxu Wang$^{1}$ \quad
Siwei Liu$^{2}$ \quad
Jinyuan Fang$^{3}$ \quad
Zaiqiao Meng$^{3}$\Thanks{~Corresponding author.} \\
$^{1}$Mohamed bin Zayed University of Artificial Intelligence \quad
$^{2}$University of Aberdeen \\
$^{3}$University of Glasgow \\
\texttt{yingxv.wang@gmail.com, siwei.liu@abdn.ac.uk} \\
\texttt{j.fang.2@research.gla.ac.uk, zaiqiao.meng@glasgow.ac.uk}
}

\def \method{EvoAgentX}

\begin{document}
\maketitle
\begin{abstract}
Multi-agent systems (MAS) have emerged as a powerful paradigm for orchestrating large language models (LLMs) and specialized tools to collaboratively address complex tasks. However, existing MAS frameworks often require manual workflow configuration and lack native support for dynamic evolution and performance optimization. In addition, many MAS optimization algorithms are not integrated into a unified framework. In this paper, we present \textbf{\method{}}, an open-source platform that automates the generation, execution, and evolutionary optimization of multi-agent workflows. \method{} employs a modular architecture consisting of five core layers: the basic components, agent, workflow, evolving, and evaluation layers. Specifically, within the evolving layer, \method{} integrates three MAS optimization algorithms, TextGrad, AFlow, and MIPRO, to iteratively refine agent prompts, tool configurations, and workflow topologies. We evaluate \method{} on HotPotQA, MBPP, and MATH for multi-hop reasoning, code generation, and mathematical problem solving, respectively, and further assess it on real-world tasks using GAIA. Experimental results show that \method{} consistently achieves significant performance improvements, including a 7.44\% increase in HotPotQA F1, a 10.00\% improvement in MBPP pass@1, a 10.00\% gain in MATH solve accuracy, and an overall accuracy improvement of up to 20.00\% on GAIA. The source code is available at: \url{https://github.com/EvoAgentX/EvoAgentX}.
\end{abstract}

\section{Introduction}

Multi-agent systems (MAS) are emerging as a powerful paradigm for orchestrating large language models (LLMs) and specialized tools to solve complex tasks collaboratively~\cite{hong2023metagpt,gao2024agentscope,fang2025comprehensive}. By coordinating multiple agents with distinct capabilities, such as planning, reasoning, or code generation, MAS decompose intricate problems into controllable subtasks and assign them to agents capable of solving them~\cite{yuan2024evoagent,zhang2025evoflow}. This flexible and modular architecture makes MAS well-suited for addressing complex real-world problems. As a result, MAS have been widely deployed in applications such as multi-hop question answering~\cite{hong2023metagpt}, software engineering automation~\cite{li2023camel}, code generation~\cite{liu2025sew}, mathematical problem solving~\cite{gao2024agentscope}, and dialogue systems~\cite{shi2024legalgpt}.

Despite these promising developments, constructing a multi-agent system typically requires manual efforts in defining the roles of agents, specifying task decomposition strategies, designing agent interactions, and configuring execution workflows~\cite{tang2024codeagent, xiao2024cellagent}. Frameworks like CrewAI~\footnote{https://www.crewai.com/}, CAMEL AI~\cite{li2023camel}, and LangGraph~\footnote{https://www.langchain.com/langgraph} have provided general-purpose frameworks for building such multi-agent systems, but they primarily rely on hand-crafted configurations or rule-based orchestration. This reliance limits scalability and usability, especially when adapting workflows to new tasks or domains. As the complexity of multi-agent systems grows, there is an urgent need for automating workflow construction to reduce manual efforts and facilitate the rapid development of agent-based applications.

Another challenge lies in enabling MAS to evolve and optimize themselves dynamically rather than relying on static, predefined configurations \cite{zhang2025evoflow}. Real-world tasks, such as multi-hop question answering or software debugging, often involve changing inputs, intermediate outcomes, or increasing task complexities \cite{xu2024generate}. These characteristics make it essential for multi-agent systems to support workflow evolution and optimization to continuously adapt strategies to meet task requirements and respond to varying conditions. By adjusting workflow topologies, modifying prompt templates, and choosing the most suitable tools, MAS can iteratively refine their workflows for improved problem-solving effectiveness \cite{zhou2024symbolic}. However, most existing MAS frameworks lack mechanisms for dynamic workflow evolution or optimization. Although some multi-agent optimization methods like DSPy~\cite{khattab2023dspy} and TextGrad~\cite{yuksekgonul2024textgrad} provide approaches for prompt refinement or agent orchestration, they remain fragmented and are not integrated into unified platforms, making it difficult for practitioners to apply and compare them consistently and effectively across diverse tasks.

\begin{figure*}[t]
    \centering
    \includegraphics[width=1.0\linewidth]{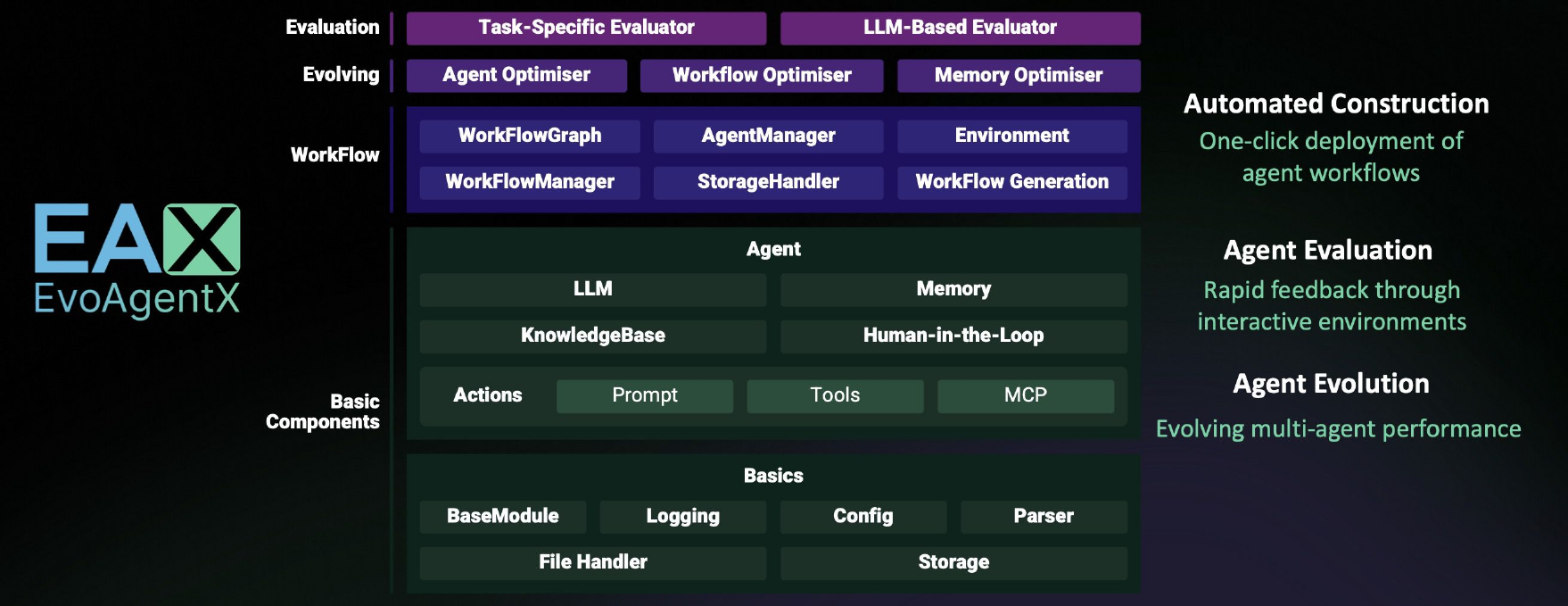}
    \caption{The framework of \method{}. It illustrates the modular architecture including the basic components, agent, workflow, evolving, and evaluation layers.}
    \label{fig:evoagentx_framework}
    \vspace{-0.5cm}
\end{figure*}

To this end, we propose \textbf{\method{}}, an open-source platform designed for the automated generation and evolutionary optimization of multi-agent workflows. As illustrated in Figure~\ref{fig:evoagentx_framework}, \method{} employs a modular architecture consisting of five core layers: basic components, agent, workflow, evolving, and evaluation. The central feature of \method{} is the evolving layer, which seamlessly integrates three state-of-the-art optimization algorithms, TextGrad~\cite{yuksekgonul2024textgrad}, AFlow~\cite{zhang2024aflow}, and MIPRO~\cite{opsahl2024optimizing}, to iteratively refine agent prompts, tool configurations, and workflow topologies. We evaluate \method{} comprehensively across diverse benchmarks, including HotPotQA for multi-hop reasoning~\cite{yang2018hotpotqa}, MBPP for code generation~\cite{austin2021program}, MATH for mathematical problem-solving~\cite{hendrycks2021measuring}, and the GAIA benchmark for real-world multi-agent tasks~\cite{mialon2023gaia}. Experimental results demonstrate that \method{} achieves substantial performance improvements through dynamic workflow evolution, including improvements of 7.44\% in HotPotQA F1, 10.00\% in MBPP pass@1, 10.00\% in MATH solve accuracy, and up to 20.00\% overall accuracy on GAIA.

The contributions of our demo are as follows:

\begin{itemize}
    \item We present \method{}, an open-source platform that enables easy customization and automatic generation of multi-agent workflows from high-level goal descriptions, reducing manual efforts in system design.
    \item \method{} integrates three optimization algorithms, TextGrad, AFlow, and MIPRO, to enable evolutionary workflow optimization.
    \item \method{} provides built-in benchmarks and standardized evaluation metrics, supporting dynamic workflow evolution that consistently improves performance on datasets such as HotPotQA, MBPP, and MATH, as well as on real-world benchmarks like GAIA.
\end{itemize}
\section{Related Work}

\subsection{Multi-Agent Systems}

Recent research on multi-agent systems (MAS) has explored how multiple language model agents can collaborate to solve complex tasks by distributing subtasks across specialized components \cite{wang2024gta, kapoor2024omniact}. Frameworks such as CAMEL AI, CrewAI, and LangGraph provide general-purpose infrastructures that allow users to define agents with distinct roles, specify communication protocols, and design execution logic and interaction patterns. These systems have enabled significant progress in areas such as multi-hop reasoning, dialogue simulation, tool-augmented reasoning, and software engineering automation \cite{ma2024agentboard}. However, most existing MAS frameworks rely on hand-crafted workflows or rule-based orchestration, requiring users to manually predefine agent hierarchies, interactions, and execution strategies for each task \cite{gao2024agentscope,islam2024mapcoder}. This places a substantial burden on users and limits the scalability and adaptability of such systems, particularly when workflows need to be adjusted for new tasks or changing conditions. To address this limitation, \method{} introduces automatic workflow generation from high-level task descriptions, eliminating the need for manual workflow design and significantly reducing human effort in building multi-agent systems.

\subsection{Multi-Agent Optimization}

Recent work on multi-agent optimization aims to enhance task performance by refining agent prompts, execution strategies, and communication structures \cite{agarwal2024promptwizard,zhou2025multi}. Early studies focused on prompt optimization, with methods such as DSPy~\cite{khattab2023dspy} and TextGrad~\cite{yuksekgonul2024textgrad} demonstrating that prompt-level refinement improves orchestration efficiency. In parallel, frameworks like DyLAN~\cite{liu2023dynamic} and Captain Agent~\cite{song2024adaptive} explored dynamic topology adjustments through reactive modifications to agent sets or communication links. A significant advancement was achieved when multi-agent topology was formalized as a search and optimization problem. Approaches such as AutoFlow~\cite{li2024autoflow} and GPTSwarm~\cite{zhuge2024gptswarm} defined topological search spaces using natural language programs or computational graphs and applied reinforcement learning to optimize agent connections. Building on this, ScoreFlow~\cite{wang2025scoreflow} and G-Designer~\cite{zhang2024g} introduced preference learning and deep generative models to produce task-adaptive communication structures. Recent work has further unified prompt and topology optimization, as in ADAS~\cite{hu2024automated}, FlowReasoner~\cite{gao2025flowreasoner}, and MaAS~\cite{zhang2025multi}, which jointly optimize agent configurations, prompts, and execution graphs using meta-search, reinforcement learning, or probabilistic supernets. Despite these advances, most existing frameworks rely on fragmented toolchains or require manual setup, making them difficult to apply consistently across diverse tasks. \method{} addresses this gap by providing an integrated platform that automates multi-agent workflow generation and unifies optimization techniques in a single end-to-end system.
\section{System Design}

\method{} is an open-source framework designed to automate the generation, execution, evaluation, and evolutionary optimization of agentic workflows. It supports automatic multi-agent workflow generation from high-level task descriptions, seamless integration of evolutionary optimization algorithms such as TextGrad~\cite{yuksekgonul2024textgrad}, AFlow~\cite{zhang2024aflow}, and MIPRO~\cite{opsahl2024optimizing}, and built-in benchmarks with standardized evaluation metrics for systematic performance assessment. To enable these capabilities, as shown in Figure~\ref{fig:evoagentx_framework}, \method{} adopts a modular architecture comprising five core layers: the basic components, agent, workflow, evolving, and evaluation layers.

\subsection{Basic Component Layer}

The basic components layer forms the foundation of \method{}, providing essential services that ensure the system's stability, scalability, and extensibility. It simplifies infrastructure management, supporting high-level agent design and workflow construction. Core modules include configuration management, logging, file handling, and storage. The configuration manager validates system parameters from structured files (e.g., YAML or JSON), while the logging module tracks system events and performance metrics. File handling components manage workflow states and agent checkpoints, ensuring experiment reproducibility. The storage manager supports both persistent and temporary storage, including caching and checkpointing.

To further enhance the system's flexibility and adaptability, \method{} integrates with a variety of LLMs through frameworks such as OpenRouter~\footnote{https://openrouter.ai/} and LiteLLM~\footnote{https://www.litellm.ai/} in the basic component layer, enabling seamless integration of LLMs from diverse sources. 

\subsection{Agent Layer}

The agent layer serves as the core functional unit of \method{}, enabling the construction of modular, intelligent entities that integrate reasoning, memory, and action execution capabilities in a seamless and flexible manner. Each agent is designed as a composition of a LLM, action modules, and memory components, together supporting flexible, context-aware decision-making and tools. 

At the center of the agent architecture, the LLM is responsible for high-level reasoning, response generation, and context interpretation. It is specified through direct instantiation or configuration files and serves as the foundation for all agent operations. Actions define the operational logic of agents. Each action encapsulates a specific task (e.g., summarization, retrieval, API invocation) and consists of a prompt template, input-output format specifications, and optional tool integrations. Formally, an agent $a_i$ is represented as:
\begin{equation}
a_i = \langle \text{LLM}_i, \text{Mem}_i, \{ \text{Act}_{i}^{(j)} \}_{j=1}^{M} \rangle,
\end{equation}
where $\text{Mem}_i$ denotes the memory module, and $\text{Act}_{i}^{(j)}$ denotes the set of action components.

An example in \ref{sec:build_agent} illustrates how to create an agent within \method{}.

\subsection{Workflow Layer}

The workflow layer is another core component of \method{}, supporting the construction, orchestration, and execution of multi-agent workflows in a structured and flexible manner. It provides a formal representation for capturing task dependencies, execution flows, and communication between agents. Each workflow is modeled as a directed graph:
\begin{equation}
\mathcal{W} = (\mathcal{V}, \mathcal{E}),
\end{equation}
where $\mathcal{V}$ denotes the set of nodes (tasks) and $\mathcal{E}$ represents directed edges encoding dependencies and data flow between tasks. Each node $v \in \mathcal{V}$ corresponds to a \texttt{WorkFlowNode}, defining a specific task, its inputs, outputs, associated agents, and execution status (\texttt{PENDING}, \texttt{RUNNING}, \texttt{COMPLETED}, or \texttt{FAILED}). Nodes can encapsulate either a set of agents, allowing dynamic selection of optimal actions during execution, or an ActionGraph specifying an explicit sequence of operations. Edges $(v_i, v_j) \in \mathcal{E}$ capture task dependencies, execution order, and optional priority weights for scheduling.

The workflow layer supports both general-purpose workflows (WorkFlowGraph) and streamlined linear workflows (SequentialWorkFlowGraph). The former provides a flexible framework for explicitly defining complex task graphs, including custom nodes, edges, conditional branches, and parallel execution patterns. It allows users to specify detailed task dependencies and exercise fine-grained control over data flow and execution logic. In contrast, the latter is designed for simplicity, automatically inferring graph connections based on task input-output dependencies and generating nodes, agents, and edges without the need for manual graph specification. This dual design facilitates rapid prototyping while preserving the expressiveness needed to model complex structures.

An example in \ref{sec:build_workflow} illustrates how to create a workflow within \method{}.

\subsection{Evolving Layer}

The evolving layer of \method{} consists of three core components: agent optimizer, workflow optimizer, and memory optimizer. These optimizers provide a unified mechanism for iteratively refining agent configurations, workflow topologies, and memory management strategies. This architecture enables the system to dynamically adapt to changing task requirements, optimize multi-agent coordination, and improve overall performance.

(1) The agent optimizer aims to refine agent prompt templates, tool configurations, and action strategies to enhance each agent's performance across diverse tasks. Formally, for an agent $a_i$ parameterized by its prompt $\text{Prompt}_i$ and configuration $\theta_i$, the optimizer seeks to compute
\begin{equation}
(\text{Prompt}_i^{(t+1)}, \theta_i^{(t+1)}) = \mathcal{O}_{\text{agent}}(\text{Prompt}_i^{(t)}, \theta_i^{(t)}, \mathcal{E}),
\end{equation}
where $\mathcal{O}_{\text{agent}}(\cdot)$ denotes the agent-level optimization operator that updates prompts and configurations based on evaluation feedback, and $\mathcal{E}$ denotes evaluation feedback. The TextGrad~\cite{yuksekgonul2024textgrad} and MIPRO~\cite{opsahl2024optimizing} optimizers are employed for agent optimization, jointly applying gradient-based prompt tuning, in-context learning, and preference-guided refinement to iteratively align prompts, tool configurations, and agent outputs with task-specific objectives based on the performance signal.

(2) The workflow optimizer focuses on improving task decomposition and execution flow by adjusting the structure of the workflow graph $\mathcal{W} = (\mathcal{V}, \mathcal{E})$. Formally, its objective is to compute
\begin{equation}
\mathcal{W}^{(t+1)} = \mathcal{O}_{\text{workflow}}(\mathcal{W}^{(t)}, \mathcal{E}),
\end{equation}
where $\mathcal{O}_{\text{workflow}}(\cdot)$ denotes the workflow-level optimization operator that updates the graph structure based on evaluation feedback, and $\mathcal{E}$ denotes evaluation feedback. The SEW~\cite{liu2025sew} and AFlow~\cite{zhang2024aflow} optimizers are employed for workflow optimization, iteratively restructuring workflow graphs by reordering nodes, modifying dependencies, and exploring alternative execution strategies guided by the task performance signal and convergence criteria.

(3) The memory optimizer remains under active development. Its objective is to provide structured, persistent memory modules $\mathcal{M}_i$ that enable selective retention, dynamic pruning, and priority-based retrieval~\cite{zeng2024structural}. Formally, it aims to compute
\begin{equation}
\mathcal{M}_i^{(t+1)} = \mathcal{O}_{\text{memory}}(\mathcal{M}_i^{(t)}, \mathcal{E}),
\end{equation}
where $\mathcal{O}_{\text{memory}}(\cdot)$ denotes the memory-level optimization operator that updates the agent’s memory module based on evaluation feedback, and $\mathcal{E}$ denotes evaluation feedback.

An example in \ref{sec:build_optimizer} demonstrates how to use the optimizer within \method{}.

\begin{figure*}[t]
    \centering
    \begin{subfigure}{0.48\linewidth}
        \centering
        \includegraphics[width=\linewidth]{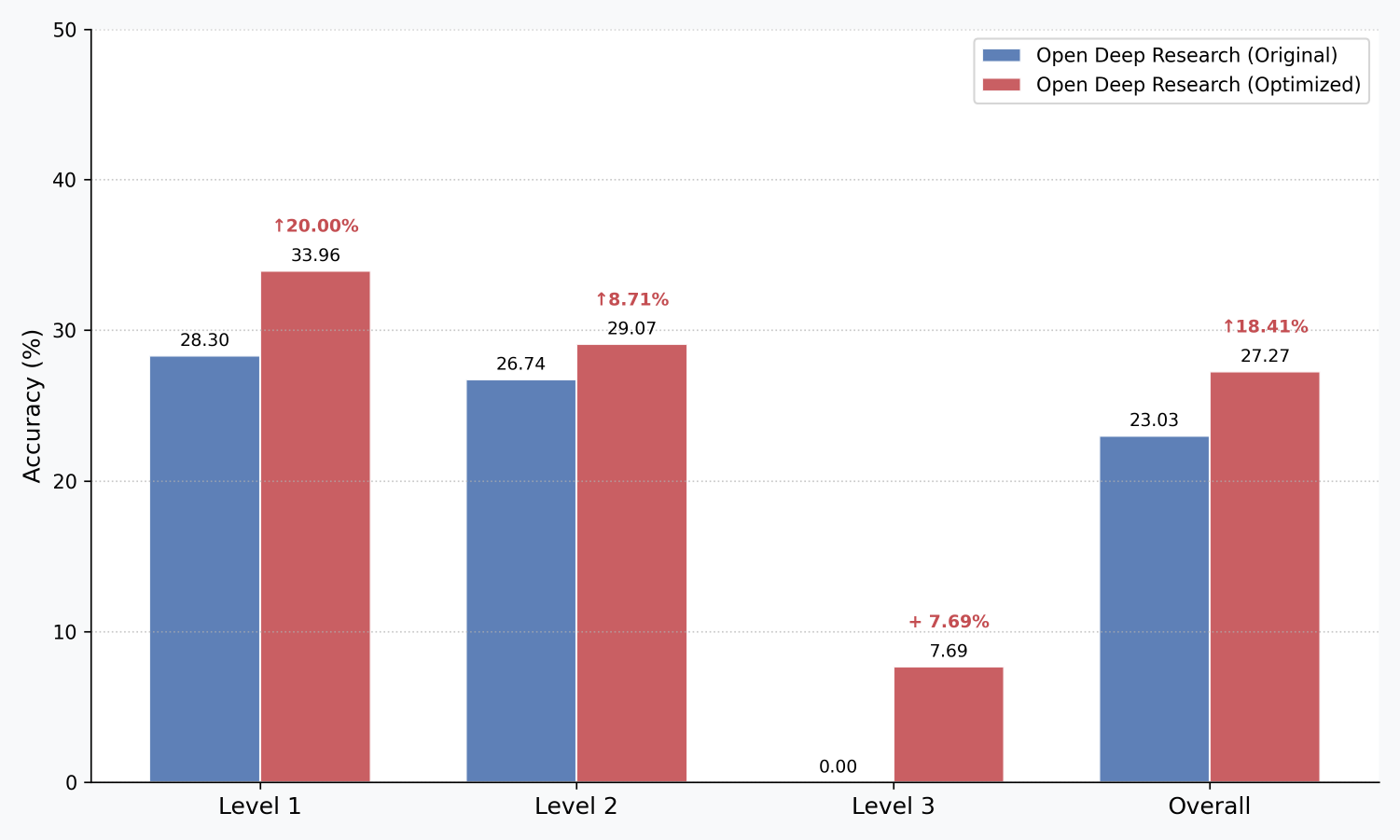}
        \caption{Performance of Open Deep Research.}
        \label{fig:open_deep_research}
    \end{subfigure}
    \hfill
    \begin{subfigure}{0.48\linewidth}
        \centering
        \includegraphics[width=\linewidth]{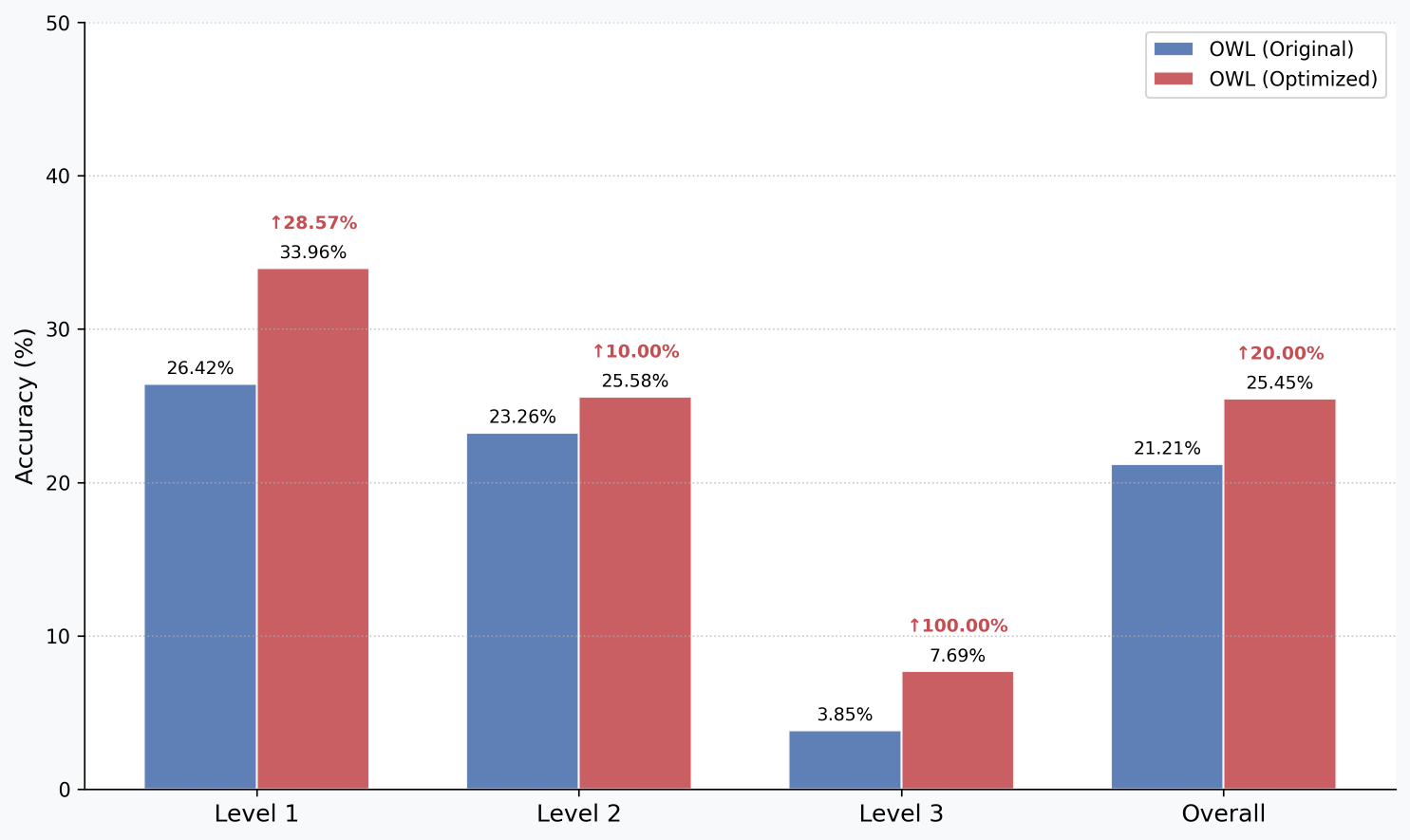}
        \caption{Performance of OWL Agent.}
        \label{fig:owl_agent}
    \end{subfigure}
    \caption{Performance improvements of Open Deep Research and OWL Agent on the GAIA benchmark.}
    \vspace{-0.5cm}
    \label{fig:gaia_results}
\end{figure*}

\begin{table}[t]
\centering
\caption{Statistics of different benchmarks. We denote Question Answering and Natural Questions as QA and NQ, respectively.}
\label{tab:dataset_stats}
\resizebox{\linewidth}{!}{
\begin{tabular}{l l r r r}
\toprule
\textbf{Task} & \textbf{Dataset Name} & \textbf{\# Train} & \textbf{\# Dev} & \textbf{\# Test} \\
\midrule
QA & NQ & 79{,}168 & 8{,}757 & 3{,}610 \\
Multi-Hop QA & HotPotQA & 90{,}447 & 7{,}405 & / \\
Math & GSM8K & 7{,}473 & / & 1{,}319 \\
Math & MATH & 7{,}500 & / & 5{,}000 \\
Code Generation & HumanEval & / & / & 164 \\
Code Generation & MBPP & / & / & 427 \\
Code Generation & LiveCodeBench (v1$\sim$v5) & / & / & 400$\sim$880 \\
Code Execution & LiveCodeBench & / & / & 479 \\
Test Output Prediction & LiveCodeBench & / & / & 442 \\
\bottomrule
\end{tabular}
}
\vspace{-0.5cm}
\end{table}

\subsection{Evaluation Layer}

The evaluation layer of \method{} provides a modular and extensible framework for systematically assessing workflow performance across tasks and benchmarks. It integrates two complementary components: (1) the task-specific evaluator and the LLM-based evaluator. The task-specific evaluator computes domain-relevant metrics by comparing workflow outputs against ground truth labels, supporting validation on datasets such as HotPotQA, MBPP, and MATH. More details about the benchmarks provided in \method{} are shown in Table~\ref{tab:dataset_stats}; (2) the LLM-based evaluator leverages the reasoning and generation capabilities of large language models to deliver flexible and context-aware evaluations. It supports qualitative assessments, consistency checking, and dynamic criteria that are not easily captured by static metrics. 

Formally, given a workflow $\mathcal{W}$ and dataset $\mathcal{D}$, the evaluation process is defined as
\begin{equation}
\mathcal{P} = \mathcal{T}(\mathcal{W}, \mathcal{D}),
\end{equation}
where $\mathcal{T}(\cdot)$ maps workflow executions to aggregated performance metrics $\mathcal{P}$. The design supports both WorkFlowGraph and ActionGraph structures, allowing evaluation at various abstraction levels.

\section{Experiments}

We evaluate \method{} across three tasks: (1) Evolution Algorithms, which optimize agent configurations and workflow topologies to improve performance; (2) Applications, where \method{} is applied to enhance multi-agent systems on real-world benchmarks; and (3) Case Study, demonstrating \method{}'s capability to optimize workflows and enhance agent performance through practical examples.

\subsection{Evolution Algorithms}

\textbf{Experimental Settings.} We evaluate \method{}’s evolutionary optimization capabilities by applying three integrated algorithmsm, TextGrad, MIPRO, and AFlow, to iteratively refine agent prompts, tool configurations, and workflow topologies, using metrics such as F1 score for multi-hop reasoning, pass@1 accuracy for code generation, and solve rate for mathematical problem solving.

\textbf{Datasets.} We assess \method{} on HotPotQA for multi-hop question answering requiring compositional reasoning~\cite{yang2018hotpotqa}, MBPP for Python code generation from natural language descriptions\cite{mbpp}, and MATH for solving high-school level mathematical problems~\cite{hendrycks2021measuring}.

\begin{table}[t]
\centering
\caption{Performance comparison across different benchmarks. \textbf{Bold} indicates the best performance.}
\label{tab:evolution_results}
\resizebox{1.0\linewidth}{!}{
\begin{tabular}{l|c|c|c}
\toprule
\multirow{2}{*}{Method} & HotPotQA & MBPP & MATH \\
 & (F1\%) & (Pass@1 \%) & (Solve \%) \\
\midrule
Original & 63.58 & 69.00 & 66.00 \\
TextGrad & \textbf{71.02} & 71.00 & \textbf{76.00} \\
AFlow & 65.09 & \textbf{79.00} & 71.00 \\
MIPRO & 69.16 & 68.00 & 72.30 \\
\bottomrule
\end{tabular}
}
\vspace{-0.5cm}
\end{table}

\textbf{Results Analysis.} As shown in Table~\ref{tab:evolution_results}, \method{}’s optimization algorithms consistently enhance performance across all benchmarks. Specifically, TextGrad substantially improves multi-hop reasoning, increasing the HotPotQA F1 score from 63.58\% to 71.02\%. AFlow significantly boosts code generation accuracy, raising MBPP pass@1 from 69.00\% to 79.00\%. Similarly, TextGrad strengthens mathematical reasoning, improving the MATH solve rate from 66.00\% to 76.00\%. These results demonstrate the \method{} can effectively refine multi-agent workflow topologies to align with task-specific objectives across domains.

\subsection{Applications}

\textbf{Experimental Settings.} We apply \method{} to optimize existing multi-agent systems on a real-world benchmark. Specifically, We select two representative open-source frameworks from the GAIA leaderboard~\cite{mialon2023gaia}, Open Deep Research~\footnote{https://github.com/langchain-ai/open\_deep\_research} and OWL~\footnote{https://github.com/camel-ai/owl}, and refine their agent prompts and workflow configurations using \method{}, evaluating performance based on accuracy, which measures the correctness of generated answers against the ground truth.

\textbf{Results Analysis.} As shown in Figure~\ref{fig:gaia_results}, \method{} significantly improves the performance of both Open Deep Research and OWL across all evaluation levels on the GAIA benchmark. For Open Deep Research, the overall accuracy increases by 18.41\%, with notable improvements of 20.00\% at Level 1, 8.71\% at Level 2, and 7.69\% at Level 3. Similarly, OWL achieves an overall accuracy improvement of 20.00\%, driven by gains of 28.57\% at Level 1, 10.00\% at Level 2, and a remarkable 100.00\% at Level 3. These results demonstrate the effectiveness of \method{} in enhancing real-world multi-agent systems through automated prompt and topology optimization.

\subsection{Case Study}

We present a case study to illustrate the practical application of \method{} in refining agent prompts and workflow configurations within existing multi-agent systems, including examples from AFlow, TextGrad, and MIPRO. More detailed analysis and results are presented in Appendix \ref{sec:appendix}.
\section{Conclusion}

We present \textbf{\method{}}, an open-source platform that automates the generation, execution, evaluation and optimization of multi-agent workflows. It addresses key limitations of existing frameworks by eliminating the need for manual workflow design and providing support for dynamic, task-specific optimization. By integrating multiple optimization algorithms, including TextGrad, AFlow, and MIPRO, \method{} enables the iterative refinement of agent prompts, tool configurations, and workflow topologies with minimal manual intervention. Experiments on diverse benchmarks, such as HotPotQA, MBPP, MATH and GAIA, demonstrate that \method{} consistently achieves substantial performance improvements in multi-hop reasoning, code generation, mathematical problem solving, and real-world multi-agent applications. In the future, we will extend \textbf{\method{}} with more optimization algorithms, richer tool integration, and long-term memory to further enhance agent adaptability and contextual awareness. We also plan to explore advanced evolution strategies~\cite{fang2025comprehensive}, including MASS~\cite{zhou2025multi}, AlphaEvolve~\cite{novikov2025alphaevolve}, and Darwin Gödel Machine~\cite{zhang2025darwin}, to advance the strategies of multi-agent optimization.

\bibliography{references}

\begin{thebibliography}{39}
\providecommand{\natexlab}[1]{#1}

\bibitem[{Agarwal et~al.(2024)Agarwal, Dani, Ganu, and Nambi}]{agarwal2024promptwizard}
Eshaan Agarwal, Vivek Dani, Tanuja Ganu, and Akshay Nambi. 2024.
\newblock Promptwizard: Task-aware agent-driven prompt optimization framework.
\newblock \emph{arXiv preprint arXiv:2405.18369}.

\bibitem[{Austin et~al.(2021{\natexlab{a}})Austin, Odena, Nye, Bosma, Michalewski, Dohan, Jiang, Cai, Terry, Le et~al.}]{austin2021program}
Jacob Austin, Augustus Odena, Maxwell Nye, Maarten Bosma, Henryk Michalewski, David Dohan, Ellen Jiang, Carrie Cai, Michael Terry, Quoc Le, and 1 others. 2021{\natexlab{a}}.
\newblock Program synthesis with large language models.
\newblock \emph{arXiv preprint arXiv:2108.07732}.

\bibitem[{Austin et~al.(2021{\natexlab{b}})Austin, Odena, Nye, Bosma, Michalewski, Dohan, Jiang, Cai, Terry, Le et~al.}]{mbpp}
Jacob Austin, Augustus Odena, Maxwell Nye, Maarten Bosma, Henryk Michalewski, David Dohan, Ellen Jiang, Carrie Cai, Michael Terry, Quoc Le, and 1 others. 2021{\natexlab{b}}.
\newblock Program synthesis with large language models.
\newblock \emph{arXiv preprint arXiv:2108.07732}.

\bibitem[{Fang et~al.(2025)Fang, Peng, Zhang, Wang, Yi, Zhang, Xu, Wu, Liu, Li et~al.}]{fang2025comprehensive}
Jinyuan Fang, Yanwen Peng, Xi~Zhang, Yingxu Wang, Xinhao Yi, Guibin Zhang, Yi~Xu, Bin Wu, Siwei Liu, Zihao Li, and 1 others. 2025.
\newblock A comprehensive survey of self-evolving ai agents: A new paradigm bridging foundation models and lifelong agentic systems.
\newblock \emph{arXiv preprint arXiv:2508.07407}.

\bibitem[{Gao et~al.(2024)Gao, Li, Pan, Kuang, Ma, Qian, Wei, Zhang, Xie, Chen et~al.}]{gao2024agentscope}
Dawei Gao, Zitao Li, Xuchen Pan, Weirui Kuang, Zhijian Ma, Bingchen Qian, Fei Wei, Wenhao Zhang, Yuexiang Xie, Daoyuan Chen, and 1 others. 2024.
\newblock Agentscope: A flexible yet robust multi-agent platform.
\newblock \emph{arXiv preprint arXiv:2402.14034}.

\bibitem[{Gao et~al.(2025)Gao, Liu, He, Dou, Du, Deng, Hooi, Lin, and Pang}]{gao2025flowreasoner}
Hongcheng Gao, Yue Liu, Yufei He, Longxu Dou, Chao Du, Zhijie Deng, Bryan Hooi, Min Lin, and Tianyu Pang. 2025.
\newblock Flowreasoner: Reinforcing query-level meta-agents.
\newblock \emph{arXiv preprint arXiv:2504.15257}.

\bibitem[{Hendrycks et~al.(2021)Hendrycks, Burns, Kadavath, Arora, Basart, Tang, Song, and Steinhardt}]{hendrycks2021measuring}
Dan Hendrycks, Collin Burns, Saurav Kadavath, Akul Arora, Steven Basart, Eric Tang, Dawn Song, and Jacob Steinhardt. 2021.
\newblock Measuring mathematical problem solving with the math dataset.
\newblock \emph{arXiv preprint arXiv:2103.03874}.

\bibitem[{Hong et~al.(2023)Hong, Zheng, Chen, Cheng, Wang, Zhang, Wang, Yau, Lin, Zhou et~al.}]{hong2023metagpt}
Sirui Hong, Xiawu Zheng, Jonathan Chen, Yuheng Cheng, Jinlin Wang, Ceyao Zhang, Zili Wang, Steven Ka~Shing Yau, Zijuan Lin, Liyang Zhou, and 1 others. 2023.
\newblock Metagpt: Meta programming for multi-agent collaborative framework.
\newblock \emph{arXiv preprint arXiv:2308.00352}.

\bibitem[{Hu et~al.(2024)Hu, Lu, and Clune}]{hu2024automated}
Shengran Hu, Cong Lu, and Jeff Clune. 2024.
\newblock Automated design of agentic systems.
\newblock \emph{arXiv preprint arXiv:2408.08435}.

\bibitem[{Islam et~al.(2024)Islam, Ali, and Parvez}]{islam2024mapcoder}
Md~Ashraful Islam, Mohammed~Eunus Ali, and Md~Rizwan Parvez. 2024.
\newblock Mapcoder: Multi-agent code generation for competitive problem solving.
\newblock In \emph{Proceedings of the 62nd Annual Meeting of the Association for Computational Linguistics (Volume 1: Long Papers)}, pages 4912--4944.

\bibitem[{Kapoor et~al.(2024)Kapoor, Butala, Russak, Koh, Kamble, AlShikh, and Salakhutdinov}]{kapoor2024omniact}
Raghav Kapoor, Yash~Parag Butala, Melisa Russak, Jing~Yu Koh, Kiran Kamble, Waseem AlShikh, and Ruslan Salakhutdinov. 2024.
\newblock Omniact: A dataset and benchmark for enabling multimodal generalist autonomous agents for desktop and web.
\newblock In \emph{European Conference on Computer Vision}, pages 161--178. Springer.

\bibitem[{Khattab et~al.(2023)Khattab, Singhvi, Maheshwari, Zhang, Santhanam, Vardhamanan, Haq, Sharma, Joshi, Moazam et~al.}]{khattab2023dspy}
Omar Khattab, Arnav Singhvi, Paridhi Maheshwari, Zhiyuan Zhang, Keshav Santhanam, Sri Vardhamanan, Saiful Haq, Ashutosh Sharma, Thomas~T Joshi, Hanna Moazam, and 1 others. 2023.
\newblock Dspy: Compiling declarative language model calls into self-improving pipelines.
\newblock \emph{arXiv preprint arXiv:2310.03714}.

\bibitem[{Li et~al.(2023)Li, Hammoud, Itani, Khizbullin, and Ghanem}]{li2023camel}
Guohao Li, Hasan Hammoud, Hani Itani, Dmitrii Khizbullin, and Bernard Ghanem. 2023.
\newblock Camel: Communicative agents for" mind" exploration of large language model society.
\newblock \emph{Advances in Neural Information Processing Systems}, 36:51991--52008.

\bibitem[{Li et~al.(2024)Li, Xu, Mei, Hua, Rama, Raheja, Wang, Zhu, and Zhang}]{li2024autoflow}
Zelong Li, Shuyuan Xu, Kai Mei, Wenyue Hua, Balaji Rama, Om~Raheja, Hao Wang, He~Zhu, and Yongfeng Zhang. 2024.
\newblock Autoflow: Automated workflow generation for large language model agents.
\newblock \emph{arXiv preprint arXiv:2407.12821}.

\bibitem[{Liu et~al.(2025)Liu, Fang, Zhou, Wang, and Meng}]{liu2025sew}
Siwei Liu, Jinyuan Fang, Han Zhou, Yingxu Wang, and Zaiqiao Meng. 2025.
\newblock Sew: Self-evolving agentic workflows for automated code generation.
\newblock \emph{arXiv preprint arXiv:2505.18646}.

\bibitem[{Liu et~al.(2023)Liu, Zhang, Li, Liu, and Yang}]{liu2023dynamic}
Zijun Liu, Yanzhe Zhang, Peng Li, Yang Liu, and Diyi Yang. 2023.
\newblock Dynamic llm-agent network: An llm-agent collaboration framework with agent team optimization.
\newblock \emph{arXiv preprint arXiv:2310.02170}.

\bibitem[{Ma et~al.(2024)Ma, Zhang, Zhu, Yang, Yang, Jin, Lan, Kong, and He}]{ma2024agentboard}
Chang Ma, Junlei Zhang, Zhihao Zhu, Cheng Yang, Yujiu Yang, Yaohui Jin, Zhenzhong Lan, Lingpeng Kong, and Junxian He. 2024.
\newblock Agentboard: An analytical evaluation board of multi-turn llm agents.
\newblock \emph{arXiv preprint arXiv:2401.13178}.

\bibitem[{Mialon et~al.(2023)Mialon, Fourrier, Wolf, LeCun, and Scialom}]{mialon2023gaia}
Gr{\'e}goire Mialon, Cl{\'e}mentine Fourrier, Thomas Wolf, Yann LeCun, and Thomas Scialom. 2023.
\newblock Gaia: a benchmark for general ai assistants.
\newblock In \emph{The Twelfth International Conference on Learning Representations}.

\bibitem[{Novikov et~al.(2025)Novikov, V{\~u}, Eisenberger, Dupont, Huang, Wagner, Shirobokov, Kozlovskii, Ruiz, Mehrabian et~al.}]{novikov2025alphaevolve}
Alexander Novikov, Ng{\^a}n V{\~u}, Marvin Eisenberger, Emilien Dupont, Po-Sen Huang, Adam~Zsolt Wagner, Sergey Shirobokov, Borislav Kozlovskii, Francisco~JR Ruiz, Abbas Mehrabian, and 1 others. 2025.
\newblock $\text{AlphaEvolve}$: A coding agent for scientific and algorithmic discovery.
\newblock \emph{arXiv preprint arXiv:2506.13131}.

\bibitem[{Opsahl-Ong et~al.(2024)Opsahl-Ong, Ryan, Purtell, Broman, Potts, Zaharia, and Khattab}]{opsahl2024optimizing}
Krista Opsahl-Ong, Michael~J Ryan, Josh Purtell, David Broman, Christopher Potts, Matei Zaharia, and Omar Khattab. 2024.
\newblock Optimizing instructions and demonstrations for multi-stage language model programs.
\newblock \emph{arXiv preprint arXiv:2406.11695}.

\bibitem[{Shi et~al.(2024)Shi, Guo, Liao, and Liang}]{shi2024legalgpt}
Juanming Shi, Qinglang Guo, Yong Liao, and Shenglin Liang. 2024.
\newblock Legalgpt: Legal chain of thought for the legal large language model multi-agent framework.
\newblock In \emph{International Conference on Intelligent Computing}, pages 25--37. Springer.

\bibitem[{Song et~al.(2024)Song, Liu, Zhang, Zhang, Luo, Wang, Wu, and Wang}]{song2024adaptive}
Linxin Song, Jiale Liu, Jieyu Zhang, Shaokun Zhang, Ao~Luo, Shijian Wang, Qingyun Wu, and Chi Wang. 2024.
\newblock Adaptive in-conversation team building for language model agents.
\newblock \emph{arXiv preprint arXiv:2405.19425}.

\bibitem[{Tang et~al.(2024)Tang, Kim, Song, Lothritz, Li, Ezzini, Tian, Klein, and Bissyand{\'e}}]{tang2024codeagent}
Xunzhu Tang, Kisub Kim, Yewei Song, Cedric Lothritz, Bei Li, Saad Ezzini, Haoye Tian, Jacques Klein, and Tegawend{\'e}~F Bissyand{\'e}. 2024.
\newblock Codeagent: Autonomous communicative agents for code review.
\newblock \emph{arXiv preprint arXiv:2402.02172}.

\bibitem[{Wang et~al.(2024)Wang, Zerun, Li, Zhang, Chen, Chen, and Le}]{wang2024gta}
Jize Wang, Ma~Zerun, Yining Li, Songyang Zhang, Cailian Chen, Kai Chen, and Xinyi Le. 2024.
\newblock Gta: a benchmark for general tool agents.
\newblock In \emph{The Thirty-eight Conference on Neural Information Processing Systems Datasets and Benchmarks Track}.

\bibitem[{Wang et~al.(2025)Wang, Yang, Li, Wang, and Aragam}]{wang2025scoreflow}
Yinjie Wang, Ling Yang, Guohao Li, Mengdi Wang, and Bryon Aragam. 2025.
\newblock Scoreflow: Mastering llm agent workflows via score-based preference optimization.
\newblock \emph{arXiv preprint arXiv:2502.04306}.

\bibitem[{Xiao et~al.(2024)Xiao, Liu, Zheng, Xie, Hao, Li, Wang, Ni, Li, Luo et~al.}]{xiao2024cellagent}
Yihang Xiao, Jinyi Liu, Yan Zheng, Xiaohan Xie, Jianye Hao, Mingzhi Li, Ruitao Wang, Fei Ni, Yuxiao Li, Jintian Luo, and 1 others. 2024.
\newblock Cellagent: An llm-driven multi-agent framework for automated single-cell data analysis.
\newblock \emph{arXiv preprint arXiv:2407.09811}.

\bibitem[{Xu et~al.(2024)Xu, He, Chen, Wang, Song, Tong, Liu, Liu, and Zhao}]{xu2024generate}
Yao Xu, Shizhu He, Jiabei Chen, Zihao Wang, Yangqiu Song, Hanghang Tong, Guang Liu, Kang Liu, and Jun Zhao. 2024.
\newblock Generate-on-graph: Treat llm as both agent and kg in incomplete knowledge graph question answering.
\newblock \emph{arXiv preprint arXiv:2404.14741}.

\bibitem[{Yang et~al.(2018)Yang, Qi, Zhang, Bengio, Cohen, Salakhutdinov, and Manning}]{yang2018hotpotqa}
Zhilin Yang, Peng Qi, Saizheng Zhang, Yoshua Bengio, William~W Cohen, Ruslan Salakhutdinov, and Christopher~D Manning. 2018.
\newblock Hotpotqa: A dataset for diverse, explainable multi-hop question answering.
\newblock \emph{arXiv preprint arXiv:1809.09600}.

\bibitem[{Yuan et~al.(2024)Yuan, Song, Chen, Tan, Li, and Yang}]{yuan2024evoagent}
Siyu Yuan, Kaitao Song, Jiangjie Chen, Xu~Tan, Dongsheng Li, and Deqing Yang. 2024.
\newblock Evoagent: Towards automatic multi-agent generation via evolutionary algorithms.
\newblock \emph{arXiv preprint arXiv:2406.14228}.

\bibitem[{Yuksekgonul et~al.(2024)Yuksekgonul, Bianchi, Boen, Liu, Huang, Guestrin, and Zou}]{yuksekgonul2024textgrad}
Mert Yuksekgonul, Federico Bianchi, Joseph Boen, Sheng Liu, Zhi Huang, Carlos Guestrin, and James Zou. 2024.
\newblock Textgrad: Automatic" differentiation" via text.
\newblock \emph{arXiv preprint arXiv:2406.07496}.

\bibitem[{Zeng et~al.(2024)Zeng, Fang, Liu, and Meng}]{zeng2024structural}
Ruihong Zeng, Jinyuan Fang, Siwei Liu, and Zaiqiao Meng. 2024.
\newblock On the structural memory of llm agents.
\newblock \emph{arXiv preprint arXiv:2412.15266}.

\bibitem[{Zhang et~al.(2025{\natexlab{a}})Zhang, Chen, Wan, Chang, Cheng, Wang, Hu, and Bai}]{zhang2025evoflow}
Guibin Zhang, Kaijie Chen, Guancheng Wan, Heng Chang, Hong Cheng, Kun Wang, Shuyue Hu, and Lei Bai. 2025{\natexlab{a}}.
\newblock Evoflow: Evolving diverse agentic workflows on the fly.
\newblock \emph{arXiv preprint arXiv:2502.07373}.

\bibitem[{Zhang et~al.(2025{\natexlab{b}})Zhang, Niu, Fang, Wang, Bai, and Wang}]{zhang2025multi}
Guibin Zhang, Luyang Niu, Junfeng Fang, Kun Wang, Lei Bai, and Xiang Wang. 2025{\natexlab{b}}.
\newblock Multi-agent architecture search via agentic supernet.
\newblock \emph{arXiv preprint arXiv:2502.04180}.

\bibitem[{Zhang et~al.(2024{\natexlab{a}})Zhang, Yue, Sun, Wan, Yu, Fang, Wang, Chen, and Cheng}]{zhang2024g}
Guibin Zhang, Yanwei Yue, Xiangguo Sun, Guancheng Wan, Miao Yu, Junfeng Fang, Kun Wang, Tianlong Chen, and Dawei Cheng. 2024{\natexlab{a}}.
\newblock G-designer: Architecting multi-agent communication topologies via graph neural networks.
\newblock \emph{arXiv preprint arXiv:2410.11782}.

\bibitem[{Zhang et~al.(2025{\natexlab{c}})Zhang, Hu, Lu, Lange, and Clune}]{zhang2025darwin}
Jenny Zhang, Shengran Hu, Cong Lu, Robert Lange, and Jeff Clune. 2025{\natexlab{c}}.
\newblock Darwin godel machine: Open-ended evolution of self-improving agents.
\newblock \emph{arXiv preprint arXiv:2505.22954}.

\bibitem[{Zhang et~al.(2024{\natexlab{b}})Zhang, Xiang, Yu, Teng, Chen, Chen, Zhuge, Cheng, Hong, Wang et~al.}]{zhang2024aflow}
Jiayi Zhang, Jinyu Xiang, Zhaoyang Yu, Fengwei Teng, Xionghui Chen, Jiaqi Chen, Mingchen Zhuge, Xin Cheng, Sirui Hong, Jinlin Wang, and 1 others. 2024{\natexlab{b}}.
\newblock Aflow: Automating agentic workflow generation.
\newblock \emph{arXiv preprint arXiv:2410.10762}.

\bibitem[{Zhou et~al.(2025)Zhou, Wan, Sun, Palangi, Iqbal, Vuli{\'c}, Korhonen, and Ar{\i}k}]{zhou2025multi}
Han Zhou, Xingchen Wan, Ruoxi Sun, Hamid Palangi, Shariq Iqbal, Ivan Vuli{\'c}, Anna Korhonen, and Sercan~{\"O} Ar{\i}k. 2025.
\newblock Multi-agent design: Optimizing agents with better prompts and topologies.
\newblock \emph{arXiv preprint arXiv:2502.02533}.

\bibitem[{Zhou et~al.(2024)Zhou, Ou, Ding, Li, Wu, Wang, Chen, Wang, Xu, Zhang et~al.}]{zhou2024symbolic}
Wangchunshu Zhou, Yixin Ou, Shengwei Ding, Long Li, Jialong Wu, Tiannan Wang, Jiamin Chen, Shuai Wang, Xiaohua Xu, Ningyu Zhang, and 1 others. 2024.
\newblock Symbolic learning enables self-evolving agents.
\newblock \emph{arXiv preprint arXiv:2406.18532}.

\bibitem[{Zhuge et~al.(2024)Zhuge, Wang, Kirsch, Faccio, Khizbullin, and Schmidhuber}]{zhuge2024gptswarm}
Mingchen Zhuge, Wenyi Wang, Louis Kirsch, Francesco Faccio, Dmitrii Khizbullin, and J{\"u}rgen Schmidhuber. 2024.
\newblock Gptswarm: Language agents as optimizable graphs.
\newblock In \emph{Forty-first International Conference on Machine Learning}.

\end{thebibliography}

\newpage
\appendix
\section{Appendix}

\subsection{Creating a Simple Agent in \method{}}\label{sec:build_agent}

In \method{}, a simple agent can be created using the \texttt{CustomizeAgent} class, which enables the quick configuration of agents with a specific prompt. To create such an agent, the first step is to configure the large language model (LLM) that will be used by the agent. This is done by defining the LLM settings, such as the model type and API key, using the \texttt{OpenAILLMConfig} class. After configuring the LLM, the agent is instantiated by specifying its name, description, and prompt, which defines the task the agent will perform. 

The following code demonstrates the process of creating and using a simple agent in \method{}, where the agent is tasked with printing "hello world." The agent is then executed, and the result is retrieved as a message object, with the content of the response extracted and displayed.

% \begin{adjustbox}{width=0.8\textwidth}
\begin{lstlisting}[language=Python, style=pythonstyle, label={lst:build_agent}, numbers=none]
from evoagentx.models import OpenAILLMConfig
from evoagentx.agents import CustomizeAgent

# Configure LLM
openai_config = OpenAILLMConfig(
    model="gpt-4o-mini",
    openai_key="YOUR_API_KEY",
    stream=True
)

# Create a simple agent
first_agent = CustomizeAgent(
    name="FirstAgent",
    description="A simple agent that prints 'hello world'",
    prompt="Print 'hello world'",
    llm_config=openai_config
)

# Execute the agent
message = first_agent()
print(f"Response from {first_agent.name}: {message.content.content}")
\end{lstlisting}
% \end{adjustbox}

This approach showcases how a basic agent can be constructed and executed with minimal configuration in \method{}.

\subsection{Creating a Simple Workflow in \method{}}\label{sec:build_workflow}

In \method{}, workflows enable multiple agents to collaborate sequentially on tasks. A basic sequential workflow involves defining tasks, each with a name, description, input-output specifications, and a prompt. To create such a workflow, the \texttt{SequentialWorkFlowGraph} class is used to define and link tasks. An agent manager oversees the agents responsible for executing each task. The workflow is executed by providing the necessary inputs, and the results are collected as outputs.

The following code illustrates the process of creating and using a simple workflow in \method{}, where the workflow includes two tasks: "Planning," in which the agent creates a detailed implementation plan for a given problem, and "Coding," in which the agent implements the solution based on the plan. The workflow is then executed with a specified problem, and the results are retrieved as outputs, with the content of each task’s result extracted and displayed sequentially.

\begin{lstlisting}[language=Python, style=pythonstyle, label={lst:build_workflow}, numbers=none]
import os
from dotenv import load_dotenv
from evoagentx.workflow import SequentialWorkFlowGraph
from evoagentx.agents import AgentManager
from evoagentx.models import OpenAILLMConfig, OpenAILLM

# Load environment variables
load_dotenv()
OPENAI_API_KEY = os.getenv("OPENAI_API_KEY")

# Configure the LLM
llm_config = OpenAILLMConfig(model=
"gpt-4o-mini", openai_key=OPENAI_API_KEY, stream=True)
llm = OpenAILLM(llm_config)

# Define tasks in the sequential workflow
tasks = [
    {
        "name": "Planning",
        "description": "Create a detailed plan for code generation",
        "inputs": [{"name": "problem", "type": "str", "required": True}],
        "outputs": [{"name": "plan", "type": "str", "required": True}],
        "prompt": "You are a software architect. Create a detailed implementation plan for the given problem.\n\nProblem: {problem}",
        "parse_mode": "str"
    },
    {
        "name": "Coding",
        "description": "Implement the code based on the plan",
        "inputs": [{"name": "problem", "type": "str", "required": True}],
        "outputs": [{"name": "code", "type": "str", "required": True}],
        "prompt": "You are a developer. Implement the code based on the provided plan.\n\nProblem: {problem}\nImplementation Plan: {plan}",
        "parse_mode": "str"
    }
]

# Create the sequential workflow graph
graph = SequentialWorkFlowGraph(goal=
"Generate code to solve programming problems", tasks=tasks)

# Initialize the agent manager and add agents
agent_manager = AgentManager()
agent_manager.add_agents_from_workflow
(graph, llm_config=llm_config)

# Create the workflow instance
workflow = WorkFlow(graph=graph, agent_manager=agent_manager, llm=llm)

# Execute the workflow with inputs
output = workflow.execute(inputs={"problem": "Write a function to find the longest palindromic substring in a given string."})

print("Workflow completed!")
print("Workflow output:\n", output)

\end{lstlisting}

This example demonstrates how a simple sequential workflow can be created and executed in \method{}, where agents collaborate in a defined sequence to accomplish a task.

\subsection{A Simple Example about using Optimizer within \method{}}\label{sec:build_optimizer}

In \method{}, the AFlow optimizer optimizes multi-agent workflows for tasks like code generation. To use the optimizer, the first step is configuring the LLMs for both optimization and execution, one for optimizing the workflow (e.g., Claude 3.5 Sonnet) and one for task execution (e.g., GPT-4o-mini). The task configuration specifies operators (e.g., Custom, CustomCodeGenerate, ScEnsemble), and the workflow is created using the \texttt{SequentialWorkFlowGraph} class. The optimizer is initialized with the paths to the workflow, LLM configurations, and optimization parameters. 

The following code demonstrates the process of creating and using an AFlow~\cite{zhang2024aflow} optimizer in \method{}, where the optimizer refines a multi-agent workflow for code generation. The optimizer is configured with specific settings, including the selection of LLMs for both optimization and execution. The process begins by configuring the \texttt{optimizer\_llm} for optimizing the workflow and the \texttt{executor\_llm} for executing the tasks. The optimizer is run with a benchmark (e.g., the HumanEval benchmark), and the results are retrieved as outputs, with each optimization step and its corresponding result displayed sequentially. 

\begin{lstlisting}[language=Python, style=pythonstyle, label={lst:optimizer}, numbers=none]
import os
from dotenv import load_dotenv
from evoagentx.optimizers import AFlowOptimizer
from evoagentx.models import LiteLLMConfig, LiteLLM, OpenAILLMConfig, OpenAILLM
from evoagentx.benchmark import AFlowHumanEval

# Load environment variables
load_dotenv()
OPENAI_API_KEY = os.getenv("OPENAI_API_KEY")
ANTHROPIC_API_KEY = os.getenv("ANTHROPIC_API_KEY")

# Configure LLMs
claude_config = LiteLLMConfig(model=
"anthropic/
claude-3-5-sonnet-20240620", anthropic_key=ANTHROPIC_API_KEY)
optimizer_llm = LiteLLM(config=claude_config)

openai_config = OpenAILLMConfig(model="gpt-4o-mini", openai_key=OPENAI_API_KEY)
executor_llm = OpenAILLM(config=openai_config)

# Initialize the benchmark
humaneval = AFlowHumanEval()

# Set up the optimizer
optimizer = AFlowOptimizer(
    graph_path=
    "examples/aflow/code_generation",  # Path to the initial workflow graph
    optimized_path=
    "examples/
    aflow/humaneval/optimized",  # Path to save optimized workflows
    optimizer_llm=optimizer_llm,  # LLM for optimization
    executor_llm=executor_llm,  # LLM for execution
    validation_rounds=3,  # Number of validation rounds
    eval_rounds=3,  # Number of evaluation rounds
    max_rounds=20,  # Maximum optimization rounds
    task_config=
    EXPERIMENTAL_CONFIG["humaneval"]  # Task configuration
)

# Optimize and test the workflow
optimizer.optimize(humaneval)
optimizer.test(humaneval)


\end{lstlisting}

This illustrates how \method{} dynamically optimizes workflows to improve performance through multiple evaluation rounds.

\subsection{Case study}\label{sec:appendix}

\subsubsection{AFlow for Workflow Optimization}

In this task, we aim to enhance the efficiency and effectiveness of multi-agent workflows for mathematical problem-solving by optimizing workflows using AFlow, as shown in the Example~\ref{lst:aflow_workflow_before} and Example~\ref{lst:aflow_workflow_after}.

The initial workflow used a basic agent configuration to solve mathematical problems with a simple prompt. After optimization with AFlow within \method{}, the workflow was significantly enhanced by incorporating detailed problem analysis, Python code generation, and solution refinement through an ensemble approach. The workflow now involves multiple agents for problem analysis, code generation, and solution refinement, improving both accuracy and clarity. This optimization results in a more comprehensive and precise problem-solving process. 

\begin{lstlisting}[language=Python, style=pythonstyle, caption={The workflow before optimization.}, label={lst:aflow_workflow_before}, numbers=none]
class Workflow:
  
   def __init__(
       self,
       name: str,
       llm_config: LLMConfig,
       benchmark: Benchmark
   ):
       self.name = name
       self.llm = create_llm_instance(llm_config)
       self.benchmark = benchmark
       self.custom = operator.Custom(self.llm)
  
   async def __call__(self, problem: str):
       
       """
       Implementation of the workflow
       """
       
       solution = await self.custom(input=problem, instruction = prompt_custom.
       SOLVE_MATH_PROBLEM_PROMPT)
       
       return solution['response']
\end{lstlisting}

\begin{lstlisting}[language=Python, style=pythonstyle, caption={The workflow after AFlow optimization.},label={lst:aflow_workflow_after}, numbers=none]
SOLVE_MATH_PROBLEM_PROMPT = r"""
Given the math problem, its analysis, and a Python code solution, provide a detailed step-by-step solution. Ensure your explanation is clear and thorough. If the code solution is relevant, incorporate its logic into your explanation.


Problem, Analysis, and Code Solution: """

FORMAT_ANSWER_PROMPT = r"""

Given the problem and its solution, extract the final numerical answer and format it in a box using LaTeX notation \boxed{}. Ensure the answer is accurate and properly formatted.

Problem and Solution: """

class Workflow:
  
   def __init__(
       self,
       name: str,
       llm_config: LLMConfig,
       benchmark: Benchmark
   ):
       self.name = name
       self.llm = create_llm_instance(llm_config)
       self.benchmark = benchmark
       self.custom = operator.Custom(self.llm)
       self.programmer = operator.Programmer(self.llm)
       self.sc_ensemble = operator.ScEnsemble(self.llm)
  
   async def __call__(self, problem: str):
       """
       Implementation of the workflow
       """
       analysis = await self.programmer(problem=problem, analysis="Analyze the math problem and provide a step-by-step approach to solve it.")
      
       code_solution = await self.programmer(problem=problem, analysis=f"Generate Python code to solve this problem: {problem}")

       solutions = []
       for _ in range(3):
           solution = await self.custom(input=problem + f"\nAnalysis: {analysis['output']}\nCode Solution: {code_solution['output']}", instruction=
           prompt_custom.
           SOLVE_MATH_PROBLEM_PROMPT)
           solutions.
           append(solution['response'])
      
       best_solution = await self.
       sc_ensemble(solutions=solutions, problem=problem)
      
       final_answer = await self.custom(input=f"Problem: {problem}\nSolution: {best_solution['response']}", instruction=
       prompt_custom.FORMAT_ANSWER_PROMPT)
      
       return final_answer['response']
\end{lstlisting}

\subsubsection{TextGrad for Prompt Optimization}

In this task, we aim to improve the efficiency and effectiveness of multi-agent workflows for mathematical problem-solving by optimizing prompts using TextGrad, as demonstrated in Example~\ref{lst:mipro_workflow_before} and Example~\ref{lst:mipro_workflow_after}.

The initial prompt employed a minimalistic instruction, directing the agent to solve mathematical problems by simply providing the final answer in a boxed format. Following optimization with TextGrad within \method{}, the prompt was substantially refined to support a structured, step-by-step reasoning process. The enhanced prompt guides the agent to assess problem complexity, apply appropriate mathematical principles, and generate executable Python code. It further emphasizes logical coherence, justification of each solution step, and the use of explanatory transitions to enhance interpretability. These improvements lead to increased solution accuracy and significantly improve the clarity and transparency of the reasoning process. 

\begin{lstlisting}[language=Python, style=pythonstyle, caption={The prompt before optimization.}, label={lst:textgrad_workflow_before}, numbers=none]
"""
Answer the math question. The answer should be in box format, e.g., \\boxed{{123}}\n
"""
\end{lstlisting}

\begin{lstlisting}[language=Python, style=pythonstyle, caption={The prompt after TextGrad optimization.},label={lst:textgrad_workflow_after}, numbers=none]
"""
Begin by assessing the complexity of the math problem to determine the appropriate level of detail required. For complex problems, provide a brief introduction to set the context and explain the relevance of key mathematical concepts. For simpler problems, focus on delivering a direct and concise solution.


Identify and apply relevant mathematical properties or theorems that can simplify the problem-solving process, such as the arithmetic sequence property. Prioritize methods that offer a concise and efficient solution, minimizing unnecessary steps while maintaining clarity.


Solve the problem using the most direct and appropriate mathematical methodologies, ensuring each calculation step is accurate. Clearly explain the reasoning behind each step, enhancing understanding by providing brief explanations of why specific mathematical properties or methods are applicable.


Maintain a smooth and coherent logical flow throughout the solution, using transitional phrases to connect different parts of the problem-solving process. Where applicable, compare alternative methods to solve the problem, discussing the benefits of each approach to provide a comprehensive understanding.


Encourage the use of visual aids, such as diagrams or charts, to illustrate complex concepts and enhance comprehension when necessary. Explicitly state and verify any assumptions made during the problem-solving process, clarifying why certain methodologies are chosen.


Conclude with a verification step to confirm the solution's correctness, and present the final answer in a consistent format, such as \\boxed{{answer}}. Ensure that the final expression is in its simplest form and that all calculations are accurate and justified.


Problem: <input>{problem}</input>


"""
\end{lstlisting}

\subsubsection{MIPRO for Prompt Optimization}

In this task, we aim to improve the efficiency and effectiveness of multi-agent workflows for mathematical problem-solving by optimizing prompts using MIPRO, as demonstrated in Example~\ref{lst:mipro_workflow_before} and Example~\ref{lst:mipro_workflow_after}.

The initial prompt was designed to provide a mathematical solution in a simple boxed format, without elaborating on the solution steps or offering detailed explanations. Following optimization with MIPRO within \method{}, the prompt was significantly improved to guide the agent through a comprehensive, step-by-step solution process. The enhanced prompt now incorporates intermediate steps, clear explanations of relevant mathematical concepts, and a thorough breakdown of the problem-solving approach. Each step is carefully articulated, ensuring a deeper understanding of the solution. This optimization not only improves the accuracy of the solution but also enhances the clarity and transparency of the reasoning process.

\begin{lstlisting}[language=Python, style=pythonstyle, caption={The prompt before optimization.}, label={lst:mipro_workflow_before}, numbers=none]
"""
Answer the math question. The answer should be in box format, e.g., \\boxed{{123}}\n
"""
\end{lstlisting}

\begin{lstlisting}[language=Python, style=pythonstyle, caption={The prompt after MIPRO optimization.},label={lst:mipro_workflow_after}, numbers=none]
"""
Please solve the following math problem, providing a detailed and clear solution process for better understanding. Ensure that the final answer is presented in a boxed format, according to the LaTeX convention (e.g., \(\boxed{{123}}\)).


Make sure to include any necessary intermediate steps, calculations, or explanations that lead you to the final answer.


**Problem:** {problem}


Examples:
1. Problem: The function $f(x)$ satisfies\n\\[f(x + y) = f(x) + f(y) + 2xy\\]for all real numbers $x$ and $y.$  If $f(1) = 4,$ then find $f(8).$
Output:
Setting $x = y,$ we get
[f(2x) = 2f(x) + 2x^2.]Then
\begin{align*}
f(2) &= 2f(1) + 2 \cdot 1^2 = 10,
f(4) &= 2f(2) + 2 \cdot 2^2 = 28,
f(8) &= 2f(4) + 2 \cdot 4^2 = \boxed{88}.
\end{align*}

2. Problem: The product $ab = 1200$, $a$ is an integer, and $b$ is an odd integer. What is the largest possible value of $b$?
Output:
Factoring out the highest power of 2 from 1200, we find that $1200=2^4\cdot75$. Therefore, the largest possible value of $b$ is $\boxed{75}$.


3. Problem: What is the product (in base 10) of the first and last digits of the base-6 representation of $682_{10}$?
Output:
We begin by converting $682_{10}$ into base-6. We see that $6^3=216$ is the largest power of 6 that is less than 682, and that $3\cdot216=648$ is the largest multiple of 216 that is less than 682. This leaves us with a remainder of $682-648=34$, which we can express as $5\cdot6^1+4\cdot6^0$. So, $682_{10}=3\cdot6^3+0\cdot6^2+
5\cdot{6^1}+4\cdot6^0=3054_6$. The first and last digits are 3 and 4, respectively, making the product of the two equal to $\boxed{12}$.


4. Problem: Compute $817_9 - 145_9 - 266_9$.  Express your answer in base 9.
Output:
$817_9 - 145_9 - 266_9 = 817_9 - (145_9 + 266_9) = 817_9 - 422_9 = \boxed{385_9}$.

"""
\end{lstlisting}

\end{document}